\documentclass[runningheads]{llncs}

\usepackage[T1]{fontenc}
\usepackage{graphicx}
\usepackage{amsmath}
\usepackage{amssymb}
\usepackage{booktabs}
\usepackage{multirow}
\usepackage{algorithm}
\usepackage{algorithmic}
\usepackage{subfigure}
\usepackage{float}
\usepackage{placeins}
\usepackage{url}
\usepackage{color}

\usepackage[colorlinks=true,linkcolor=blue,citecolor=blue,urlcolor=blue,filecolor=blue]{hyperref}
\usepackage[table]{xcolor}
\usepackage{tcolorbox}

\begin{document}

\title{E-SDS -- Environment-aware See it, Do it, Sorted: Automated Environment-Aware Reinforcement Learning for Humanoid Locomotion}

\titlerunning{E-SDS: Environment-Aware See it, Do it, Sorted}

\author{Enis Yalcin \and
        Joshua O'Hara \and
        Maria Stamatopoulou\and
        Chengxu Zhou\and
        Dimitrios Kanoulas
}

\authorrunning{Enis Yalcin et al.}

\institute{University College London, Department of Computer Science, London, UK\\
\email{\{enis.yalcin.24, joshua.o'hara.24, maria.stamatopoulou.21, chengxu.zhou, d.kanoulas\}@ucl.ac.uk}}

\maketitle

\begin{abstract}
Vision-language models (VLMs) show promise in automating reward design in humanoid locomotion, which could eliminate the need for tedious manual engineering. However, current VLM-based methods are essentially ``blind'', as they lack the environmental perception required to navigate complex terrain. We present E-SDS (Environment-aware See it, Do it, Sorted), a framework that closes this perception gap. E-SDS integrates VLMs with real-time terrain sensor analysis to automatically generate reward functions that facilitate training of robust perceptive locomotion policies, grounded by example videos. Evaluated on a Unitree G1 humanoid across four distinct terrains (simple, gaps, obstacles, stairs), E-SDS uniquely enabled successful stair descent, while policies trained with manually-designed rewards or a non-perceptive automated baseline were unable to complete the task. In all terrains, E-SDS also reduced velocity tracking error by 51.9–82.6\%. Our framework reduces the human effort of reward design from days to less than two hours while simultaneously producing more robust and capable locomotion policies.

\end{abstract}

\section{Introduction}
Deep reinforcement learning (RL) has become a primary method for the development of dynamic locomotion controllers for humanoid robots~\cite{Radosavovic2023}. However, a significant bottleneck in this process is the reliance on manual reward engineering. Designing effective reward functions is a time-consuming and brittle process, often requiring the careful tuning of numerous weighted terms to elicit a single desired behavior~\cite{VanMarum2024}. This manual effort limits the scalability of controller development and the diversity of skills that can be practically acquired.

To address this challenge, two distinct research directions have emerged. The first focuses on automated reward generation, where vision-language models (VLMs) are used to synthesize reward functions from high-level instructions or video demonstrations~\cite{Stamatopoulou2024}. Although these methods successfully automate the engineering process, the policies they train are fundamentally blind. They operate on proprioceptive state or video-derived motion priors, lacking any awareness of the robot's immediate physical environment. Consequently, their application has been limited to locomotion on simple, regular terrains. The second direction is perceptive locomotion, where exteroceptive sensors such as height scanners and LiDAR are integrated to enable navigation in complex, unstructured environments~\cite{Long2024}. These approaches have proven that environmental awareness is critical for robust performance, but still rely on manual rewards to translate sensory data into desired actions.

Thus, a fundamental gap exists: existing work does not unify automated reward generation with environment-aware control. The policies derived from automated systems cannot perceive the terrain, and the perceptive systems cannot automate their reward design. We introduce E-SDS (Environment-aware See it, Do it, Sorted) to bridge this gap. E-SDS is a framework for training perceptive locomotion policies via RL that conditions VLM-based reward synthesis directly on real-time terrain statistics, making the reward generation process itself environment-aware. Our contributions are: (1) A novel framework that automatically generates reward functions for perceptive humanoid locomotion conditioned on quantitative terrain statistics, such as gap ratios and obstacle densities. (2) An iterative refinement process that uses training feedback to systematically eliminate failure modes and improve the robustness of policy without human intervention. (3) A demonstration that this approach generates policies capable of complex behaviors, such as stair descent, that are unattainable by automated methods, either manual or perception-blind. E-SDS reduces the velocity tracking error by 51.9--82.6\% compared to a manually tuned baseline and generates robust policies in less than two hours, a process that typically requires days of expert engineering.

\section{Related Work}
Manual reward engineering remains a primary obstacle to the implementation of reinforcement learning (RL) in locomotion, as minor changes in weights or terms can induce unintended behaviors and fragile policies~\cite{VanMarum2024}. In legged robotics, policy optimization is commonly performed with PPO, often on scale with massively parallel simulation, and has recently produced humanoid locomotion capable in the real world~\cite{Radosavovic2023}. To reduce the burden of reward specification, a growing body of work uses learned or generated rewards. Human preference learning replaces hand-crafted objectives with a reward model learned from pairwise trajectory comparisons~\cite{Christiano2017}. Vision-language models (VLMs) have been used as zero-shot reward models that map a natural-language description to a task-conditioned evaluator~\cite{Rocamonde2024}. Closer to our setting are the methods that synthesize executable reward code. Eureka employs an LLM code to propose and iteratively refine reward functions, enabling non-trivial skills in simulation and on hardware~\cite{Ma2023}. Text2Reward maps natural language to symbolic reward programs for downstream RL~\cite{Xie2023}. SDS~\cite{Stamatopoulou2024} extends this direction by requiring a VLM with a grid of video frames and structured skill analyzes to generate compact reward functions for quadrupedal gait imitation from a single demonstration, followed by closed-loop refinement during training. These approaches reduce human effort, but operate without explicit access to the robot’s environment state during reward generation, which limits their application on unorganized or cluttered terrain.

A separate line of work shows that exteroception is crucial for high performance in complex terrain. Robust perceptive locomotion integrates onboard elevation maps or depth sensing with proprioception and learns policies that adapt footfall and body posture in advance of contact. Rapid Motor Adaptation augments a blind base policy with an adaptive module trained with privileged information, improving transfer and tolerance to terrain variation~\cite{Kumar2021RMA}. Similar integration of perception and control has been demonstrated in navigation~\cite{Shanks2025DreamerNav}, locomotion~\cite{Raghavan2020Agile}, and manipulation~\cite{Beddow2024RLGrasping}, where the coupling of sensorimotor feedback and learning improves robustness and generalization. Despite these advances, the reward specification for perceptive locomotion typically remains manual and task specific~\cite{VanMarum2024}, which restricts scalability across terrains and platforms.

E-SDS lies at the intersection of automated reward generation and perceptive locomotion. Compared to previous automated methods~\cite{Stamatopoulou2024}, the E-SDS conditions reward the synthesis of terrain statistics computed from robot sensors, which steers the generated code toward terms that explicitly leverage height scans and LiDAR. Compared to perception-driven locomotion~\cite{Zhou2024Teleoperation}~\cite{Wang2025}, E-SDS removes most manual reward tuning by closing the loop between environment analysis, code generation, and training feedback. This environment-aware automation enables perceptive humanoid policies that handle stairs, gaps, and clutter, addressing a gap between perception-blind automated rewards and manually engineered perceptive controllers.

\section{Methodology}
\begin{figure}[t]
    \centering
    \includegraphics[width=\textwidth]{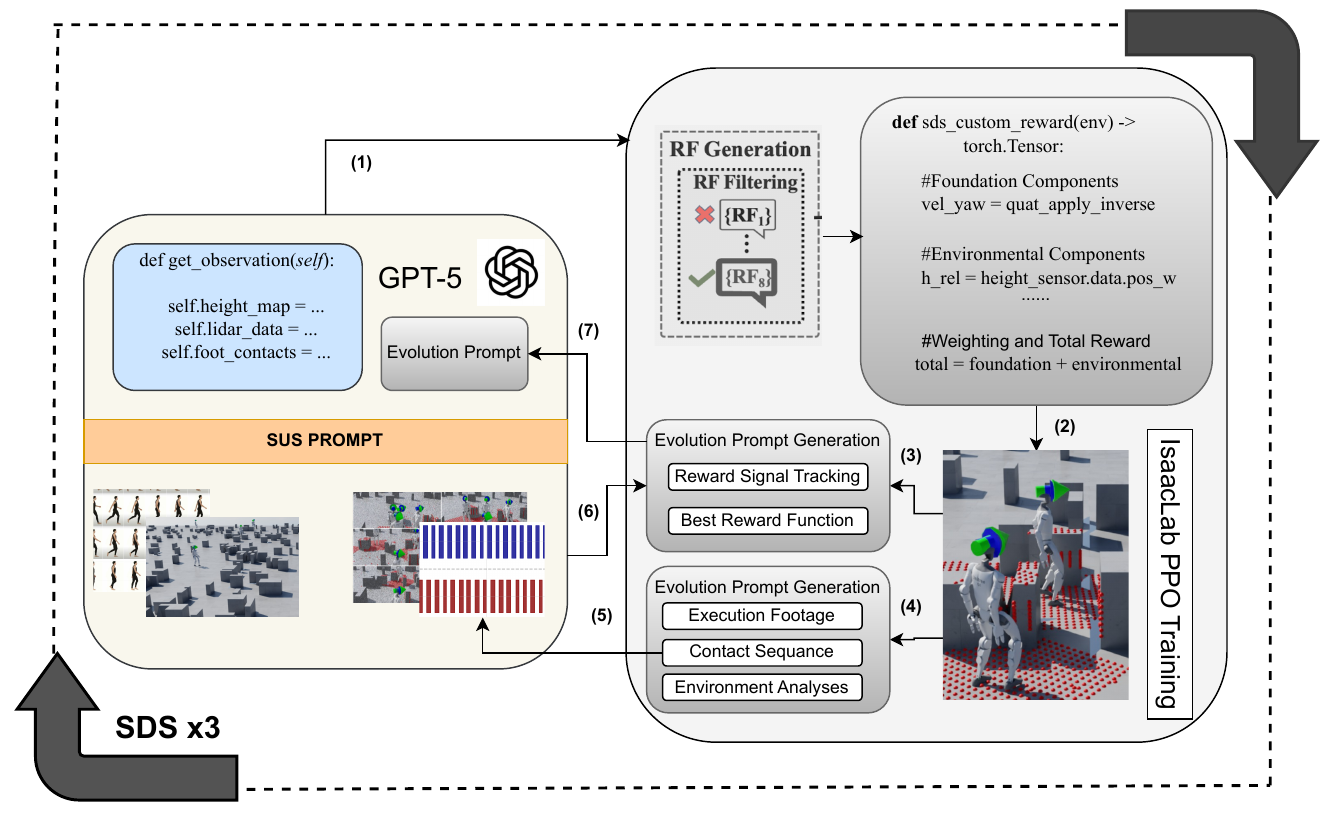}
    \caption{E-SDS pipeline showing the automated reward generation and refinement.}
    \label{fig:pipeline}
\end{figure}

The E-SDS framework automates the generation of environment-aware reward functions to train perceptive humanoid locomotion policies. The system operates through a closed-loop pipeline that combines multi-agent VLM analysis of video demonstrations with quantitative terrain analysis (Fig.~\ref{fig:pipeline}). The use of a Python-based RL framework, NVIDIA Isaac Lab, allows us to reframe the challenge of defining the reward function $R$ as a conditional code generation problem, which can be addressed by a large language model.

\subsection{Problem Formulation}
We formulate perceptive humanoid locomotion as a partially observable Markov decision process (POMDP), which we address with a recurrent policy that implicitly handles the state history. The objective is to learn a policy $\pi(a_t|s_t, h_t)$ that maps an observation $s_t \in \mathcal{S}$ and a hidden state $h_t$ to an action $a_t \in \mathcal{A}$. The action $a_t$ corresponds to a 23-dimensional vector of target joint torques for the Unitree G1 humanoid. Crucially, observation $s_t$ is a high-dimensional vector (792 dimensions) comprising both proprioceptive state and exteroceptive environmental data. The proprioceptive component includes joint positions, velocities, and base orientation. The exteroceptive component consists of processed data from an onboard height scanner ($27 \times 21$ grid) and a LiDAR sensor (144 measurements), providing the policy with direct perception of the immediate terrain. The policy is trained through reinforcement learning to maximize the expected cumulative discounted reward $\mathbb{E}\left[\sum_{t=0}^{T} \gamma^t R(s_t, a_t)\right]$. E-SDS automates the synthesis of $R$ by conditioning its generation on both the desired skill and the specific environment.

\subsection{Environment-Aware Reward Generation Agent}
E-SDS generates reward functions using a multi-agent system built upon a vision-language model (GPT-5). This process extends the foundation of SDS~\cite{Stamatopoulou2024} by incorporating a novel agent dedicated to environmental analysis. The process begins by encoding the video demonstration for VLM comprehension using \textbf{Grid-Frame Prompting}. The video is adaptively sampled into a sequence of frames, which are arranged into a single composite image grid. This preserves spatio-temporal information while being efficient for VLM processing. This grid is then analyzed using the \textbf{SUS (See it, Understand it, Sorted)} prompting strategy, a multi-stage, chain-of-thought approach where specialized agents deconstruct the desired behavior. A \textit{Contact Sequence analyzer} and \textit{Gait analyzer} extract footfall patterns, while a \textit{Task Requirement analyzer} identifies high-level objectives such as target velocity and posture. The core innovation of E-SDS is the introduction of an \textit{Environment Analysis Agent}, which provides quantitative context on the physical terrain. This agent first deploys a fleet of 1000 robots in the target simulation environment for a brief period (10 seconds) to collect sensor data. Then it processes these data to compute a statistical summary, including obstacle density, gap ratios, and terrain roughness. This statistical summary is integrated with the behavioral description from the SUS analysis. The combined environment-sensitive prompt is then passed to the VLM for code-generation to synthesize a Python reward function (Algorithm~\ref{alg:reward_generation}). This function includes not only terms that encourage the desired gait, but also environment-specific terms that leverage the robot's onboard sensors.

\begin{algorithm}
\caption{Environment-Aware Reward Function Generation}
\label{alg:reward_generation}
\begin{algorithmic}[1]
\STATE \textbf{Input:} Video demonstration $V$, Environment configuration $E$
\STATE \textbf{Output:} Executable Python reward function $R$
\STATE \textit{// Video Analysis}
\STATE $I_{grid} \leftarrow \text{GridFramePrompting}(V)$
\STATE $P_{skill} \leftarrow \text{SUS\_Analysis}(I_{grid})$ \COMMENT{Extracts gait, contacts, task goals}
\STATE \textit{// Environment Analysis}
\STATE $D_{terrain} \leftarrow \text{CollectSensorData}(E, \text{num\_robots}=1000)$
\STATE $S_{terrain} \leftarrow \text{ComputeStatistics}(D_{terrain})$ \COMMENT{Gap ratio, obstacle density, etc.}
\STATE \textit{// Code Synthesis}
\STATE $P_{combined} \leftarrow \text{CombinePrompts}(P_{skill}, S_{terrain})$
\STATE $R \leftarrow \text{VLM.GenerateCode}(P_{combined})$
\STATE \textbf{return} $R$
\end{algorithmic}
\end{algorithm}

\subsection{The E-SDS Training and Refinement Pipeline}
The agent reward function is used within a closed-loop training and refinement pipeline to produce the final perceptive policy (Algorithm~\ref{alg:esds_pipeline}). This process is fully automated. Let $i$ be the current iteration index, starting at $i=0$.

\textbf{1. Reward Synthesis.} At the start of iteration $i$, the reward generation agent (Section 3.2) is prompted to generate a set of candidate reward functions $N=2$, $\{R_k^{(i)}\}_{k=1}^N$.

\textbf{2. Perceptive Policy Training.} For each candidate reward function $R_k^{(i)}$, a corresponding perceptive policy $\pi_k^{(i)}$ is trained for $T=500$ iterations. Training is performed using Proximal Policy Optimization (PPO) in a massively parallel simulation environment with 3000 robot instances. The PPO objective is:
\begin{equation}
L^{\text{TOTAL}}(\theta) = \hat{\mathbb{E}}_t\left[L^{\text{CLIP}}(\theta) - c_1 L^{VF}(\theta) + c_2 S[\pi_\theta](s_t)\right]
\end{equation}
where $L^{\text{CLIP}}$ is the clipped surrogate objective, $L^{VF}$ is the value function loss, and $S$ is an entropy bonus.

\textbf{3. Automated Evaluation.} After training, each policy $\pi_k^{(i)}$ is evaluated. The system collects quantitative performance metrics, such as velocity tracking error and torso contact rate, and captures rollout footage. A \textit{Feedback Agent} uses the VLM to analyze these data and assign a performance score $J(\pi_k^{(i)})$.

\textbf{4. Iterative Reward Refinement.} The best-performing reward function of the current iteration, $R^{*(i)} = \arg\max_{R_k^{(i)}} J(\pi_k^{(i)})$, is selected. The code for $R^{*(i)}$, along with natural-language feedback describing the policy's performance (e.g. identifying failure modes such as freezing), is used to seed the prompt for the next iteration, $i+1$. This closed-loop process is repeated for a total of three iterations, progressively refining the reward function. The entire pipeline completes in approximately 99 minutes for each terrain.

\begin{algorithm}
\caption{The E-SDS Training and Refinement Pipeline}
\label{alg:esds_pipeline}
\begin{algorithmic}[1]
\STATE \textbf{Input:} Video demonstration $V$, Environment configuration $E$
\STATE \textbf{Initialize:} Initial prompt $P^{(0)}$, number of iterations $I_{max}=3$
\FOR{$i=0$ \TO $I_{max}-1$}
    \STATE \textit{// 1. Reward Synthesis}
    \STATE $\{R_k^{(i)}\}_{k=1}^N \leftarrow \text{RewardGenerationAgent}(P^{(i)})$
    \FOR{$k=1$ \TO $N$}
        \STATE \textit{// 2. Policy Training}
        \STATE $\pi_k^{(i)} \leftarrow \text{TrainPolicy}(R_k^{(i)}, \text{iterations}=T)$
        \STATE \textit{// 3. Automated Evaluation}
        \STATE $J(\pi_k^{(i)}), F_k \leftarrow \text{EvaluatePolicy}(\pi_k^{(i)})$ \COMMENT{$F_k$ is feedback text}
    \ENDFOR
    \STATE \textit{// 4. Iterative Refinement}
    \STATE $k^* \leftarrow \arg\max_k J(\pi_k^{(i)})$
    \STATE $R^{*(i)} \leftarrow R_{k^*}^{(i)}$
    \STATE $P^{(i+1)} \leftarrow \text{UpdatePrompt}(R^{*(i)}, F_{k^*})$
\ENDFOR
\STATE \textbf{Output:} Final refined policy $\pi_{k^*}^{(I_{max}-1)}$
\end{algorithmic}
\end{algorithm}

\section{Experiments \& Results}
We conducted a series of experiments to evaluate the E-SDS framework against relevant baselines. The evaluation was designed to: (1) compare the performance of our automated, environment-aware approach against a manually-engineered perceptive policy; (2) analyze the different locomotion strategies that emerge on complex terrains; and (3) perform a direct ablation to quantify the importance of environment awareness in the automated reward generation process.

\begin{figure}[htbp]
\centering
\includegraphics[width=0.7\textwidth]{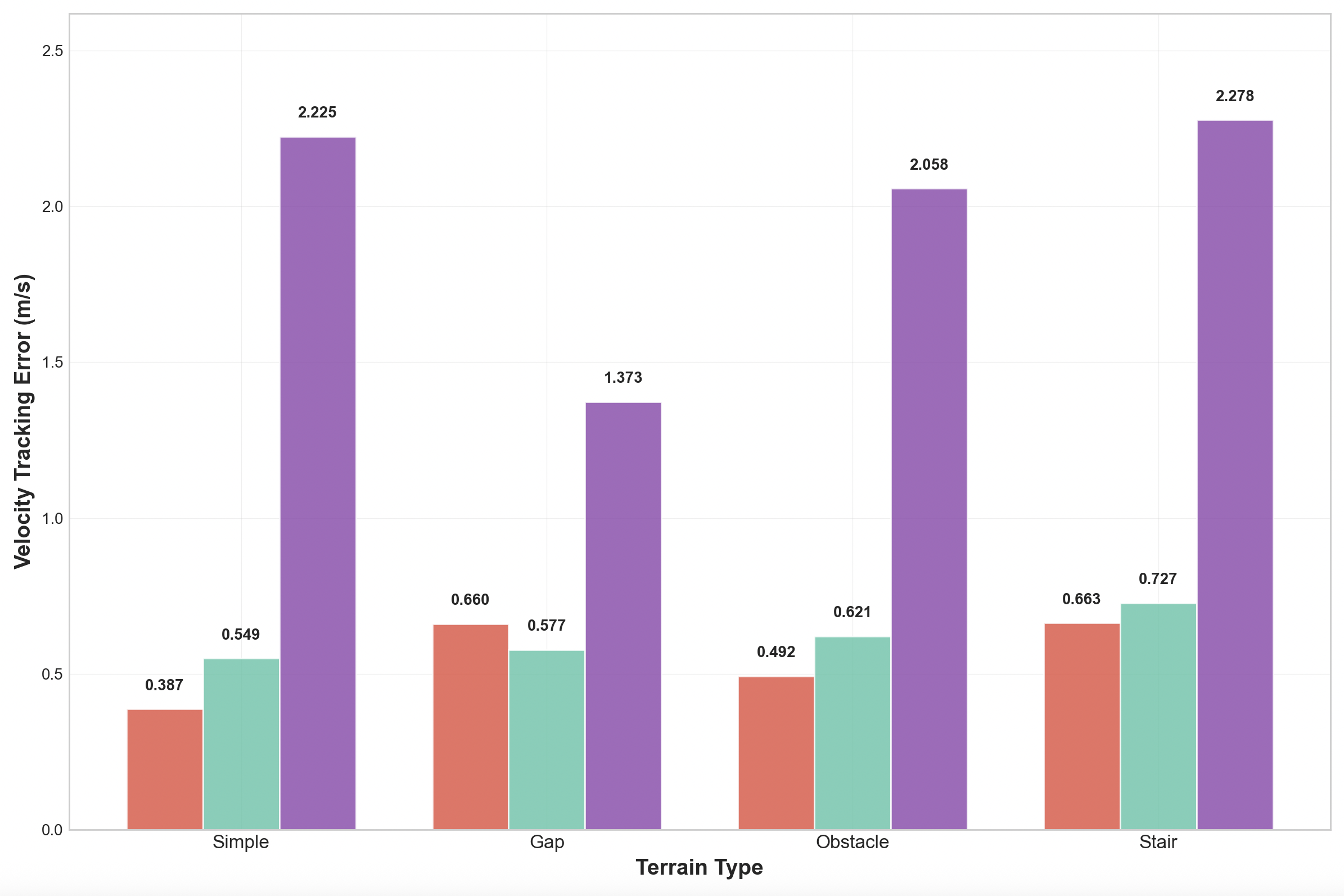}
\caption{Velocity tracking error between E-SDS (red), Foundation (green), Baseline (Purple).}
\label{fig:vte}
\end{figure}

\subsection{Experimental Setup}

\textbf{Environments.} All policies were evaluated on a simulated Unitree G1 humanoid robot across four distinct terrains of increasing complexity: a \textit{simple} terrain with gentle bumps (3-5cm); a \textit{gaps} terrain with sections of missing ground (80-120cm width); an \textit{obstacles} terrain cluttered with blocks of varying sizes; and a \textit{stairs} terrain requiring controlled descent down 12cm steps.

\textbf{Baselines for Comparison.} We compare the performance of three policies:
\begin{itemize}
    \item \textbf{E-SDS (Environment-Aware):} Our proposed method, where both the reward generation process and the final policy have access to environmental sensor data (height scanner and LiDAR).
    \item \textbf{Foundation-Only:} A direct ablation of our core contribution. This policy is trained using a reward function generated without environmental analysis, and the policy itself operates on proprioceptive data only.
    \item \textbf{Baseline (Manual Rewards):} A perceptive policy trained using 13 manually designed and tuned reward terms from existing literature~\cite{VanMarum2024}. This policy has the same sensor access as E-SDS.
\end{itemize}

\textbf{Metrics.} Performance is assessed using a suite of quantitative metrics. Velocity tracking error measures command following accuracy, defined as the Euclidean distance between the achieved and commanded velocities:
\begin{equation}
e_{\text{vel}} = \sqrt{(v_x - v^{\text{cmd}}_x)^2 + (v_y - v^{\text{cmd}}_y)^2 + (\omega_z - \omega^{\text{cmd}}_z)^2}
\end{equation}
The exploration score quantifies area coverage by rewarding visits to new grid cells ($N_{\text{cells}}$), distance from origin ($R_{\text{max}}$), and recent movement ($\Delta d$):
\begin{equation}
E_{\text{score}} = 0.5 \cdot N_{\text{cells}} + 2.0 \cdot R_{\text{max}} + \min(10 \cdot \Delta d, 5.0)
\end{equation}
Safety is measured primarily by the rate of contact with the torso, which counts the number of falls per episode. Locomotion quality assesses the naturalness and smoothness of the gait on a scale of [0, 1].

\subsection{Comparative Performance Analysis}
The results show that E-SDS generates policies that outperform the manually-engineered baseline, with the performance gap widening as terrain complexity increases.

\textbf{Stair Descent.} The stairs provide the clearest differentiation between methods. Only the E-SDS policy successfully learned to descend the stairs, achieving a high exploration score (10.93) and zero torso contacts (Table~\ref{tab:stair_results}). The manual rewards baseline policy, despite having sensor access, remained stationary at the top of the stairs, resulting in a low exploration score (3.50). The perception-blind Foundation-Only policy attempted to walk forward, leading to a high rate of falls (torso contact rate of 333.5). This shows that sensor access alone is insufficient; the reward function must be intelligently structured to take advantage of sensory information, a task that E-SDS successfully automates.

\begin{table}[htbp]
\caption{Stair Terrain Performance Comparison}
\label{tab:stair_results}
\centering
\small
\begin{tabular}{@{}lccc@{}}
\toprule
\textbf{Metric} & \textbf{Env-Aware} & \textbf{Found-Only} & \textbf{Baseline} \\
\midrule
Locomotion Quality & \textbf{0.412} & 0.342 & 0.393 \\
Exploration Score & \textbf{10.930} & 11.109 & 3.495 \\
Torso Contact Rate & \textbf{0.000} & 333.466 & \textbf{0.000} \\
Velocity Tracking (m/s) & \textbf{0.663} & 0.727 & 2.278 \\
\bottomrule
\end{tabular}
\end{table}

\textbf{Navigation of Discontinuous and Cluttered Terrains.} On the Gap and Obstacle terrains, E-SDS learns an \textit{active navigation strategy}, while the Baseline adopts a \textit{conservative avoidance strategy}. As shown in Tables \ref{tab:gap_results} and \ref{tab:obstacle_results}, E-SDS explores 2.07x more area on the gap terrain and 2.36x more area on the obstacle terrain than the Baseline. The Baseline achieves a low torso contact rate on gaps by using its height sensors to maintain a constant elevation, thereby avoiding the gaps rather than navigating between them. Despite obstacles, it remains largely stationary. In contrast, E-SDS actively navigates both environments to maximize area coverage.

\begin{table}[htbp]
\caption{Gap Terrain Performance Comparison}
\label{tab:gap_results}
\centering
\small
\begin{tabular}{@{}lccc@{}}
\toprule
\textbf{Metric} & \textbf{Env-Aware} & \textbf{Found-Only} & \textbf{Baseline} \\
\midrule
Velocity Tracking (m/s) & 0.660 & \textbf{0.577} & 1.373 \\
Exploration Score & \textbf{10.886} & 5.170 & 6.447 \\
Torso Contact Rate & 5.136 & 140.476 & \textbf{1.492} \\
\bottomrule
\end{tabular}
\end{table}

\begin{table}[htbp]
\caption{Obstacle Terrain Performance Comparison}
\label{tab:obstacle_results}
\centering
\small
\begin{tabular}{@{}lccc@{}}
\toprule
\textbf{Metric} & \textbf{Env-Aware} & \textbf{Found-Only} & \textbf{Baseline} \\
\midrule
Velocity Tracking (m/s) & \textbf{0.492} & 0.621 & 2.058 \\
Exploration Score & \textbf{7.825} & 5.873 & 5.870 \\
Torso Contact Rate & \textbf{37.92} & 316.98 & 46.13 \\
\bottomrule
\end{tabular}
\end{table}

\textbf{Locomotion on Simple Terrain.} Even on the simplest terrain, E-SDS demonstrates a clear advantage in command following. While maintaining comparable stability and safety (zero torso contacts), the E-SDS policy achieves a velocity tracking error of 0.387 m/s, an 82.6\% improvement over the Baseline's error of 2.225 m/s (Table~\ref{tab:simple_results}). The Baseline's poor tracking performance is correlated with it remaining stationary 6.2\% of the time, compared to just 0.3\% for E-SDS.

\begin{table}[htbp]
\caption{Simple Terrain Performance Comparison}
\label{tab:simple_results}
\centering
\small
\begin{tabular}{@{}lccc@{}}
\toprule
\textbf{Metric} & \textbf{Env-Aware} & \textbf{Found-Only} & \textbf{Baseline} \\
\midrule
Velocity Tracking (m/s) & \textbf{0.387} & 0.549 & 2.225 \\
Exploration Score & \textbf{6.898} & 6.541 & 6.895 \\
Torso Contact Rate & \textbf{0.000} & 1.584 & \textbf{0.000} \\
\bottomrule
\end{tabular}
\end{table}

\subsection{Ablation Study: The Necessity of Environment Awareness}
The comparison between the E-SDS (Environment-Aware) and Foundation-Only policies serves as a direct ablation of the environment-aware components of our framework. The results confirm that integrating environmental analysis into the automated reward generation pipeline is essential to develop competent policies in complex terrains. Without perception, the performance of the Foundation-Only policy degrades significantly. In the gap terrain, its torso contact rate of 140.5 is 27.4 times higher than that of the environment-aware policy (5.1). This is a direct result of the policy being unable to perceive and avoid gaps, leading to frequent falls. A similar trend is observed on the obstacle terrain, where the Foundation-Only policy's torso contact rate of 317.0 is 8.4 times higher than the 37.9 achieved by the environment-aware policy. The necessity of environmental awareness is most pronounced in the stair terrain. The Foundation-Only policy, unable to perceive the steps, completely fails, registering a torso contact rate of 333.5. In contrast, the environment-aware E-SDS policy navigates the same terrain with zero falls. These results demonstrate that conditioning the automated reward generation process on environmental statistics is a critical factor in the system's success, enabling it to synthesize rewards that produce robust, perceptive behaviors.

\section{Discussion}

\begin{figure}[htbp]
\centering
\includegraphics[width=0.8\textwidth]{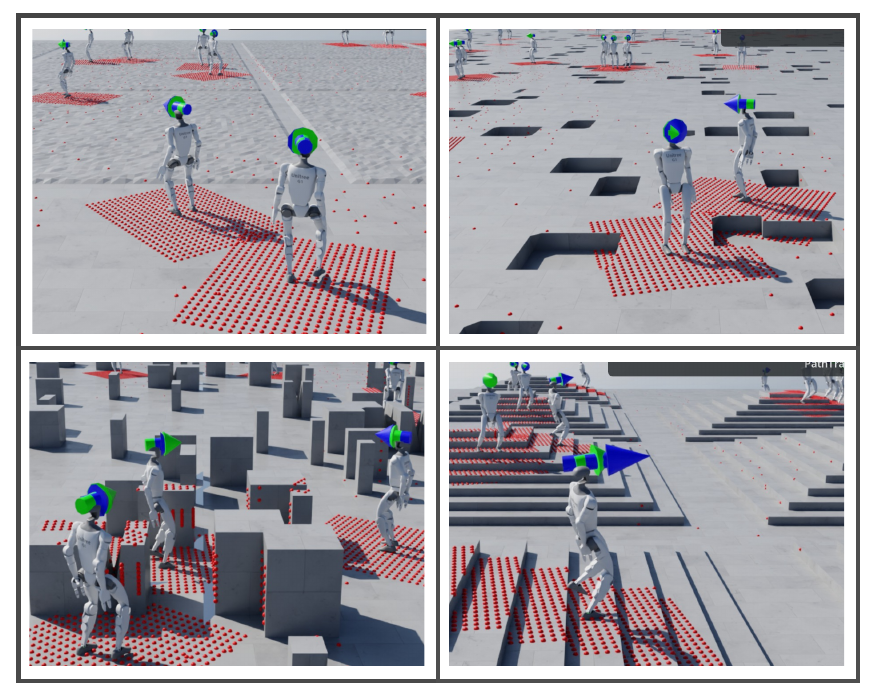}
\caption{Evaluation tasks in Isaac Lab. Simple (top left), gap (top right), obstacle (bottom left), stairs (bottom right).}
\label{fig:task}
\end{figure}

Our experimental evaluation demonstrates that conditioning automated reward generation on environment statistics produces policies that are more capable than both perception-blind automated methods and manually engineered perceptive baselines. The results show that including environmental perception is crucial for dealing with complex terrains with obstacles, gaps, and uneven surfaces. The automated E-SDS framework consistently outperformed a manually tuned baseline, reducing velocity tracking error by 51.9-82.6\% in approximately 99 minutes per terrain, a fraction of the time and expertise required for manual engineering. The framework's closed-loop iterative refinement was also vital for achieving robustness, automatically identifying and correcting emergent failure modes, such as 'freezing' behavior on gap terrains. Despite its strong performance, the framework has key limitations that inform future work. Its current approach of generating a specialized policy for each terrain does not scale to mixed, real-world environments, pointing to the need for a unified, multi-task policy. Furthermore, the evaluation was conducted exclusively in simulation, making the transfer to physical hardware by bridging the sim-to-real gap a critical next step. Lastly, while the refinement process is automated, the initial setup still relies on manual prompt engineering.

\section{Conclusion}
In this work, we presented E-SDS, a framework to address the challenge of manual reward engineering for perceptive humanoid locomotion. By conditioning automated reward generation on quantitative terrain statistics, E-SDS bridges the gap between perception-blind automated systems and perceptive but manually-tuned policies. Our experiments demonstrated that E-SDS generates policies that significantly outperform a manually engineered baseline, reducing the velocity tracking error by 51.9-82.6\% and uniquely enabling complex skills such as stair descent, where other methods failed. The framework generates these more capable policies in under two hours, a fraction of the time required for manual design. By establishing that environment conditioning is essential for automated reward generation in complex terrains, this work represents a significant step toward creating more autonomous robots capable of learning robust skills in unstructured environments.

\begin{credits}
\subsubsection{\ackname} This work was partially supported by UKRI FLF [MR/V025333/1] (RoboHike). For Open Access, the author has applied a CC BY copyright license to any manuscript version arising from this submission.
\end{credits}

\bibliographystyle{splncs04}
\bibliography{references}
\end{document}